\documentclass[runningheads]{llncs}
\usepackage{graphicx}
\usepackage[colorlinks,linkcolor=red]{hyperref}
\usepackage{tikz}
\usepackage{comment} 
\usepackage{amsmath,amssymb} 
\usepackage{color}
\usepackage{multirow}

\begin{document}
\pagestyle{headings}
\mainmatter
\def\ECCVSubNumber{3128}  

\title{Improving Multispectral Pedestrian Detection by Addressing Modality Imbalance Problems} 

\titlerunning{ Addressing Modality Imbalance Problems}
%
\author{Kailai Zhou\inst{1} \and
Linsen Chen\inst{1}\and
Xun Cao\inst{1}}
\authorrunning{Kailai Zhou, Linsen Chen, Xun Cao }
%
\institute{Nanjing University, Nanjing, China \\
\email{\{calayzhou,linsen\}@smail.nju.edu.cn \{caoxun\}@nju.edu.cn}}
\maketitle

\begin{abstract}
Multispectral pedestrian detection is capable of adapting to insufficient illumination conditions by leveraging color-thermal modalities. On the other hand, it is still lacking of in-depth insights on how to fuse the two modalities effectively. Compared with traditional pedestrian detection, we find multispectral pedestrian detection suffers from modality imbalance problems which will hinder the optimization process of dual-modality network and depress the performance of detector. Inspired by this observation, we propose Modality Balance Network (MBNet) which facilitates the optimization process in a much more flexible and balanced manner. Firstly, we design a novel Differential Modality Aware Fusion (DMAF) module to make the two modalities complement each other. Secondly, an illumination aware feature alignment module selects complementary features according to the illumination conditions and aligns the two modality features adaptively. Extensive experimental results demonstrate MBNet outperforms the state-of-the-arts on  both the challenging KAIST and CVC-14 multispectral pedestrian datasets in terms of the accuracy and the computational efficiency. Code is available at \url{https://github.com/CalayZhou/MBNet}.
\keywords{Multispectral pedestrian detection  $\cdot$ Modality imbalance problems $\cdot$ Multimodal feature fusion}
\end{abstract}

\section{Introduction}

Recent years have witnessed increasing researches towards object detection among vision community by taking the advantages of multi-modal inputs, such as RGB + thermal, RGB + depth, RGB + LiDAR and so on \cite{ju2014depth,hwang2015multispectral,ha2017mfnet,behley2019semantickitti}. Compared than traditional single-modal RGB images, which present great challenges at complex scenarios (e.g. dim environment, face spoofing detection \cite{zhang2019casia}, autonomous driving \cite{li2016unified,wu2017squeezedet}, etc), the introducing of another modality dramatically benefits the tasks of object detection. For instances, spectral images are able to detect the optical radiation of matter and reveal the essential color properties of target object, to avoid the metamerism ambiguity. Thermal images can be captured based on the heat radiation difference of the object, which does not rely on external light sources. Time-of-flight (TOF) or LiDAR sensors provide additional depth information of the target scene, which has been widely used as data representation for many vision applications. Even with these remarkable benefits, however, how to effectively fuse multi-modal information in the context of advanced algorithms, like convolutional neural network, still remains much to be studied.

\begin{figure}[t]
	\begin{center}
		\includegraphics[width=0.9\linewidth]{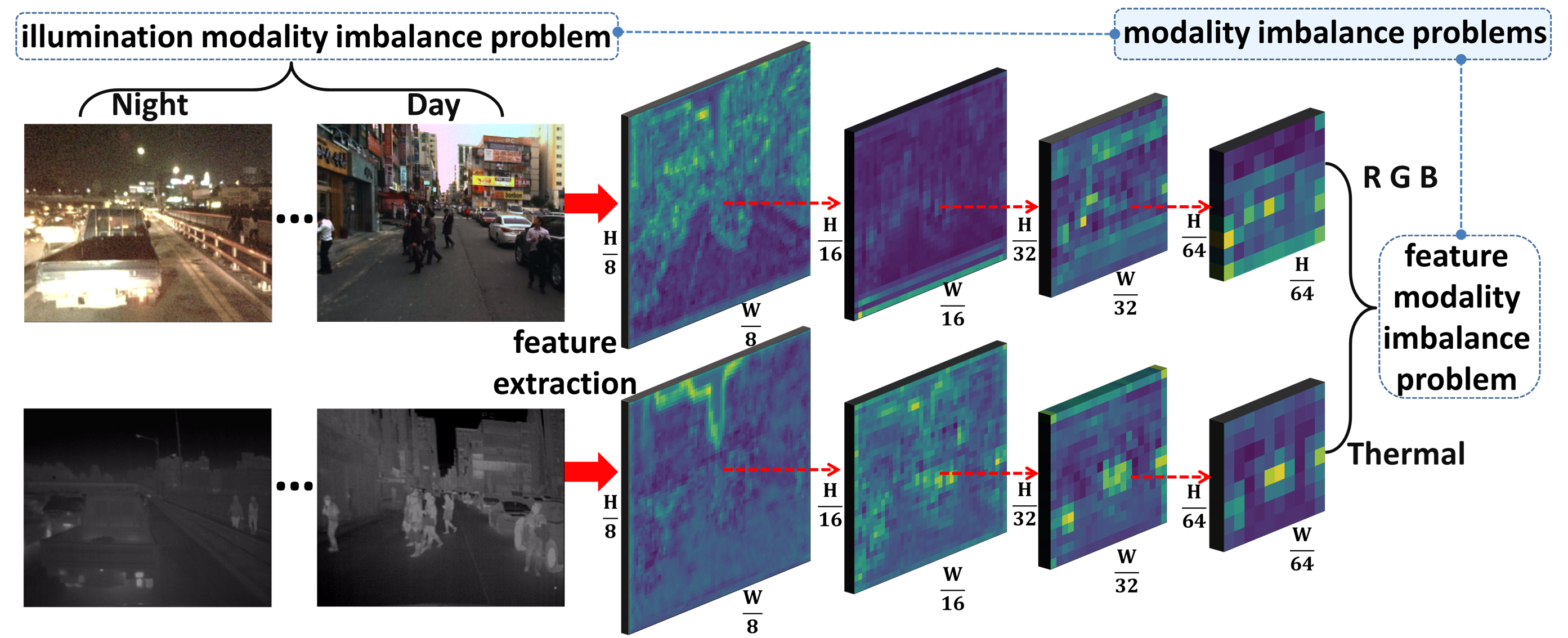}
	\end{center}
	\caption{The modality imbalance problems which consist of two parts: the illumination modality imbalance problem and the feature modality imbalance problem.}
	
	\label{fig:modality}
\end{figure}

As for ordinary optimization process of object detection from multi-modality inputs, the imbalance problems \cite{oksuz2019imbalance} are crucial. The most known imbalance problem is the foreground-to-background imbalance \cite{lin2017focal}. This drawback is caused by an extremely inequality between the number of positive examples and negative ones. Nevertheless, the imbalance problems are not limited to the class imbalance. For instance, in multi-task losses minimization, the imbalance problems exist since the norms of gradients are different and the ranges of loss functions vary \cite{guo2018dynamic}. The common solution is to add coefficients upon each loss function to guide a balanced optimization process. Similarly, the modality imbalance issue in multispectral detection has a substantial influence on the algorithm performance.

The traditional Caltech \cite{dollar2011pedestrian} and CityPersons \cite{zhang2017citypersons} pedestrian detection datasets only have RGB modality images captured during the day, so as shown in Fig.~\ref{fig:modality}, modality imbalance problems existing in multispectral pedestrian detection datasets can be divided into two categories: the illumination modality imbalance and the feature modality imbalance. Illumination modality imbalance means the difference of illumination conditions between the daytime and the night images. Intuitively, pedestrians in RGB images have clearer texture features than thermal images in daytime. Comparatively, thermal images can provide more distinct pedestrian shapes than RGB images during night time. The RGB modality branch and the thermal modality branch tend to obtain different confidence scores and have uneven contributions to the object losses under diverse illumination conditions. It is expected that the RGB modality branch and the thermal modality branch should be optimized adaptively according to illumination conditions \cite{cao2019box,li2019illumination}. 

Feature modality imbalance problem signifies that the misalignment and inadequate integration of different modalities can lead to an uneven contribution and representation of the features. On the one hand, as the visualization results shown in Fig.~\ref{fig:modality}, it is obvious that the RGB and thermal modality features are diverse in terms of pedestrian morphology, texture and properties in the two independent backbone networks. In RGB modality, the complexion and hair of the pedestrian can be some important hints of the pedestrian characteristics \cite{chi2019relational}, but none of the thermal images has such cues. It is necessary to sufficiently incorporate the cross-modality complementarity to generate robust features. On the other hand, the misalignment between the RGB and thermal modalities will cause unbalanced modality feature representation in the fixed receptive fields of a convolution kernel. Both the balance and the integration of different modalities are the cornerstone we should consider in multispectral pedestrian detection. Unfortunately, existing RGB-Thermal based detection methods simply fuse the RGB and the thermal input/features by concatenation \cite{li2019illumination,zhang2019cross1,li2018multispectral,wagner2016multispectral}. The inherent complementary is not fully exploited yet between different modalities.

To address the modality imbalance problems above, we investigate the impact and explore solutions in this paper. First, we construct the Modality Balance Network (MBNet) based on SSD \cite{liu2016ssd} to extract the characteristics of two modalities separately. Then for the purpose of fully fusing features at different scales in the network,   Differential Modality Aware Fusion (DMAF) module is proposed to tap the difference between RGB and thermal feature maps which brings more complementary information at each channel. Finally we design an illumination aware feature alignment module to align two modality features and induce the network to be optimized adaptively according to illumination conditions. 

The main contributions of this paper are as follows: (1) We present modality imbalance problems specific to multispectral pedestrian detection, and analyse that modality imbalance problems will affect the performance of the detector due to the modality inconsistency in the optimization of the network; (2) We propose a one-stage detector named Modality Balance Network (MBNet) which consists of Differential Modality Aware Fusion (DMAF) module and  illumination aware feature alignment (IAFA) module to address the modality imbalance problems. With DMAF module and IAFA module, the contribution of each feature map from two modalities will be explicitly integrated and balanced. In addition, The MBNet backbone (ResNet embedded with DMAF) may also do a favor to other computer vision communities; (3) MBNet achieves state-of-the-art results on both the challenging KAIST and CVC-14 multispectral pedestrian datasets in terms of the accuracy while maintaining the fastest speed.

\section{Relate Work}

\subsection{Multispectral Pedestrian Detection}
CNN-based pedestrian detection has achieved notable progress in recent years with methods of occlusion handling \cite{zhang2018occluded,noh2018improving,wang2018repulsion}, cascaded detection systems \cite{liu2018learning,brazil2019pedestrian}, semantic attention \cite{zhou2019ssa,brazil2017illuminating}, anchor-free approach \cite{liu2019high}, etc. Nevertheless, current pedestrian detectors using single RGB modality may fail under the insufficient illumination condition.The KAIST multispectral pedestrian detection dataset \cite{hwang2015multispectral} provides a new way to solve this problem by combining RGB modality and thermal modality. The initial baseline F + T + THOG is extended from Aggregated Channel Features (ACF) \cite{dollar2014fast} with the thermal channel added. As the popularization of deep learning, the CNN-based methods \cite{wagner2016multispectral,choi2016multi,park2018unified,xu2017learning} greatly reduce the miss rate of multispectral pedestrian detection. Inspired by \cite{zhang2016faster}, Boosted Decision Trees classifier \cite{konig2017fully} is built on high-resolution RPN feature maps to reduce potential false positive detections. MSDS RCNN \cite{li2018multispectral} is learned by jointly optimizing pedestrian detection and semantic segmentation tasks.

How to fuse the information of two modalities is the common concerned problem in multispectral pedestrian detection. Liu et al. \cite{liu2016multispectral} design four distinct fusion architectures that integrate two modality branches on different DNNs stages and reveal the Halfway Fusion model provides the best performance. GFD-SSD \cite{zheng2019gfd} proposes two variations of novel Gated Fusion Units (GFU) that learn the combination of feature maps generated by the two SSD middle layers. Zhang et al. \cite{zhang2019cross1} explore the cross-modality disparity problem in multispectral pedestrian detection and propose a novel region feature alignment module to solve this problem. CIAN \cite{zhang2019cross0} makes the middle-level feature maps of two streams converge to a unified one under the guidance of cross-modality interactive attention and adopts the context enhancement blocks (CEBs) to further augment contextual information. Illumination-aware Faster R-CNN \cite{li2019illumination} adaptively merges color and thermal sub-networks to obtain the final confidence scores via a gate function defined over the illumination value.
As the most popular solution, the two-stream architecture with concatenating RGB-Thermal feature maps has achieved significant improvements. Nevertheless, direct concatenation will inevitably introduce redundant features and a selection module is required to unveil the relation of modality complementary features. 

\subsection{Imbalance Problems In Object Detection}
Oksuz et al. \cite{oksuz2019imbalance} present a comprehensive review of the imbalance problems in object detection and group these problems in a taxonomic tree with four main types: spatial imbalance, objective imbalance, class imbalance and scale imbalance. Spatial imbalance and objective imbalance focus on spatial properties of the bounding boxes and multiple loss functions respectively. Class imbalance occurs due to the significant inequality among different classes of training data. RetinaNet \cite{lin2017focal} addresses class imbalance by means of reshaping the standard cross entropy loss to prevent the vast number of easy negatives from overwhelming the detector. AP-Loss \cite{chen2019towards} and DR Loss \cite{qian2019dr} also provide ideas of designing loss function to solve the class imbalance problem. Scale imbalance occurs when certain sizes of the object bounding boxes are over-represented in the network. For instances, SSD \cite{liu2016ssd} makes independent predictions from features at different layers. Since abstractness of information varies among different layers, it is unreliable to make predictions directly from different layers of the backbone network. Feature Pyramid Network \cite{lin2017feature} exploits an additional top-down pathway in order to have a balanced mixed of features from different scales. FPN can be further enhanced \cite{liu2018path} by integrating and refining pyramidal feature maps.

In addition to the integration balance of different level, we argue that the integration of different modality features should also be balanced in the two-stream network. In other words, different modality features should be fully integrated and represented in order to have a balanced modality optimization in the training.

\begin{figure}[t]
	\begin{center}
		\includegraphics[width=1.0\linewidth]{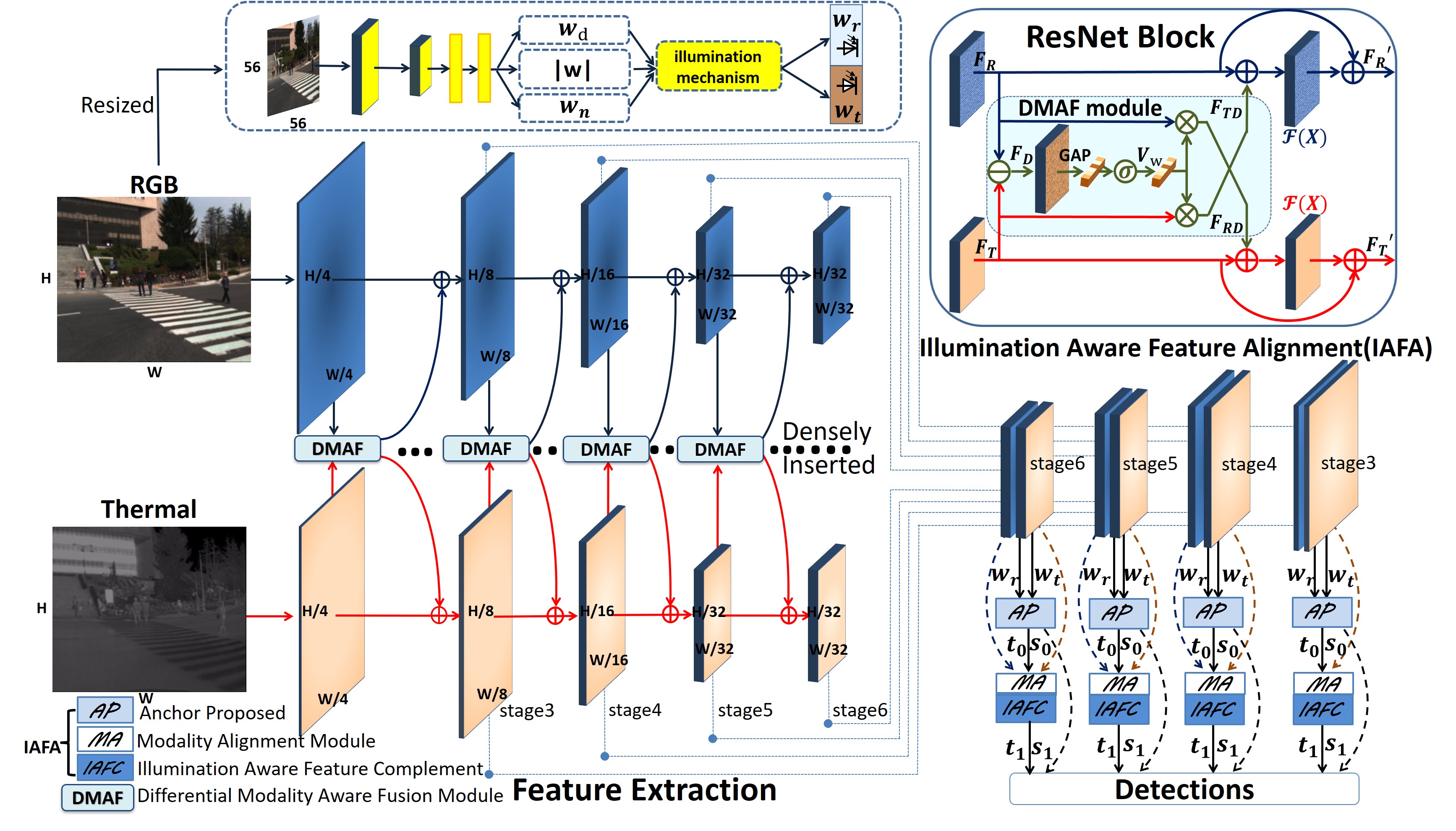}
	\end{center}
	\caption{Overview framework of the Modality Balance Network (MBNet). The MBNet consists of three parts: feature extraction module, illumination aware feature alignment module and illumination mechanism. The feature extraction module adopts ResNet-50 \cite{He_2016_CVPR} as the backbone network and embeds DMAF module to supplement modality information. Illumination mechanism is designed to acquire illumination values which will assign weights to two modality streams. Illumination aware feature alignment module plays the role of adapting the model to different illumination conditions and aligning the two modality features in the region proposal stage. }
	\label{fig:framework}
\end{figure}

\section{Approach}
The overall architecture of the proposed method is shown in Fig.~\ref{fig:framework}. The MBNet extends the framework of SSD \cite{liu2016ssd} and it consists of three parts: feature extraction module, illumination aware feature alignment module and illumination mechanism. Details of DMAF module are introduced in Sec.~\ref{sec:DMAF_sec}. The design of illumination aware feature alignment module is introduced in Sec.~\ref{sec:IAM}.


\subsection{Differential Modality Aware Fusion Module}
\label{sec:DMAF_sec} 

To address feature modality imbalance problem, we propose to enhance the one modality from another modality with differential modality information. Previous RGB-T fusion models \cite{wagner2016multispectral,li2018multispectral,zhang2019cross0,zhang2019cross1} based on deep convolutional networks typically employ a two-stream architecture, in which the RGB and thermal modalities are learned independently. The most straightforward method is to concatenate the features at different levels, e.g., early fusion, halfway fusion as well as late fusion \cite{liu2016multispectral,konig2017fully,li2019illumination}. However, it is ambiguous to capture the cross-modality complementary information by traditional direct concatenation scheme. Both modalities have their own characteristic representations which are mixed with useful hints and noises. While simple fusion strategies such as linear combination or concatenation are lacking in clarity to extract cross-modality complementary. In our view, the inherent difference between the two modalities can be exploited with an explicit and simple mechanism named Differential Modality Aware Fusion (DMAF) module.

We are inspired by differential amplifier circuits in which the common-mode signals are suppressed and the differential-mode signals are amplified. Our DMAF module retains the original features and compensates according to differential features. The RGB convolution feature map $ F_{R}$ and the thermal convolution feature map $ F_{T}$ can be represented with common modality part and differential modality part at each channel as follows: 
\begin{equation}
\begin{split}	
F_{T} &=\frac{F_{T}+F_{T}}{2}+\frac{F_{R}-F_{R}}{2}=\frac{F_{R}+F_{T}}{2}+\frac{F_{T}-F_{R}}{2} \\ 
F_{R} &=\frac{F_{R}+F_{R}}{2}+\frac{F_{T}-F_{T}}{2}=\frac{F_{R}+F_{T}}{2}+\frac{F_{R}-F_{T}}{2}
\end{split}
\end{equation}

\noindent where the common modality part reflects the common features and the differential modality part reflects the unique features captured by two modalities. Eq. 1 illustrates the principle of splitting which is same behind differential amplifier circuits and DMAF module.
The key idea of our DMAF module is acquiring complementary features from another modality with channel-wise differential weighting. We expect the learning of complementary features to be enhanced by explicitly modeling modality interdependencies, so that the network sensitivity to informative features from another modality can be increased. 

In order to make sufficient use of cross-modality complements, the DMAF module is densely inserted in each ResNet block. As the top right corner of Fig.~\ref{fig:framework} shows, we obtain the differential feature $F_{D}$ by direct subtraction of two modalities first. Then we squeeze global spatial information $F_{D}$ into a global differential vector which contains channel-wise differential statistics with global average pooling. The global differential vector can be interpreted as a channel descriptor whose statistics are expressive for the discrepancy between RGB and thermal modality. The tanh activation function ranging from -1 to 1 is applied for the global differential vector to obtain the fusion weight vector $V_{w}$. The two modality features $F_{T}$ and $F_{R}$  are recalibrated by the fusion weight vector $V_{w}$ with channel-wise multiplication. The recalibration results $F_{RD}$,$F_{TD}$ will be added to the original modality path as complementary information. After the enhancement from another modality with DMAF module, the more informative and robust features are generated and sent to the next ResNet block in the following step. The whole procedure of DMAF module can be formulated as:
\begin{equation}
\begin{aligned} 
F_{T}^{\prime} &=F_{T}+ \mathcal{F}\left(F_{T} \oplus F_{RD}\right) 
= F_{T}+\mathcal{F}\left(F_{T} \oplus\left(\sigma\left(G A P\left(F_{D}\right)\right) \odot F_{R}\right)\right) \\
F_{R}^{\prime} &=F_{R}+ \mathcal{F}\left(F_{R} \oplus F_{TD}\right) = F_{R}+\mathcal{F}\left(F_{R} \oplus\left(\sigma\left(G A P\left(F_{D}\right)\right) \odot F_{T}\right)\right)
\end{aligned}
\end{equation}

\begin{figure}[t]
	\begin{center}
		\includegraphics[width=0.6\linewidth]{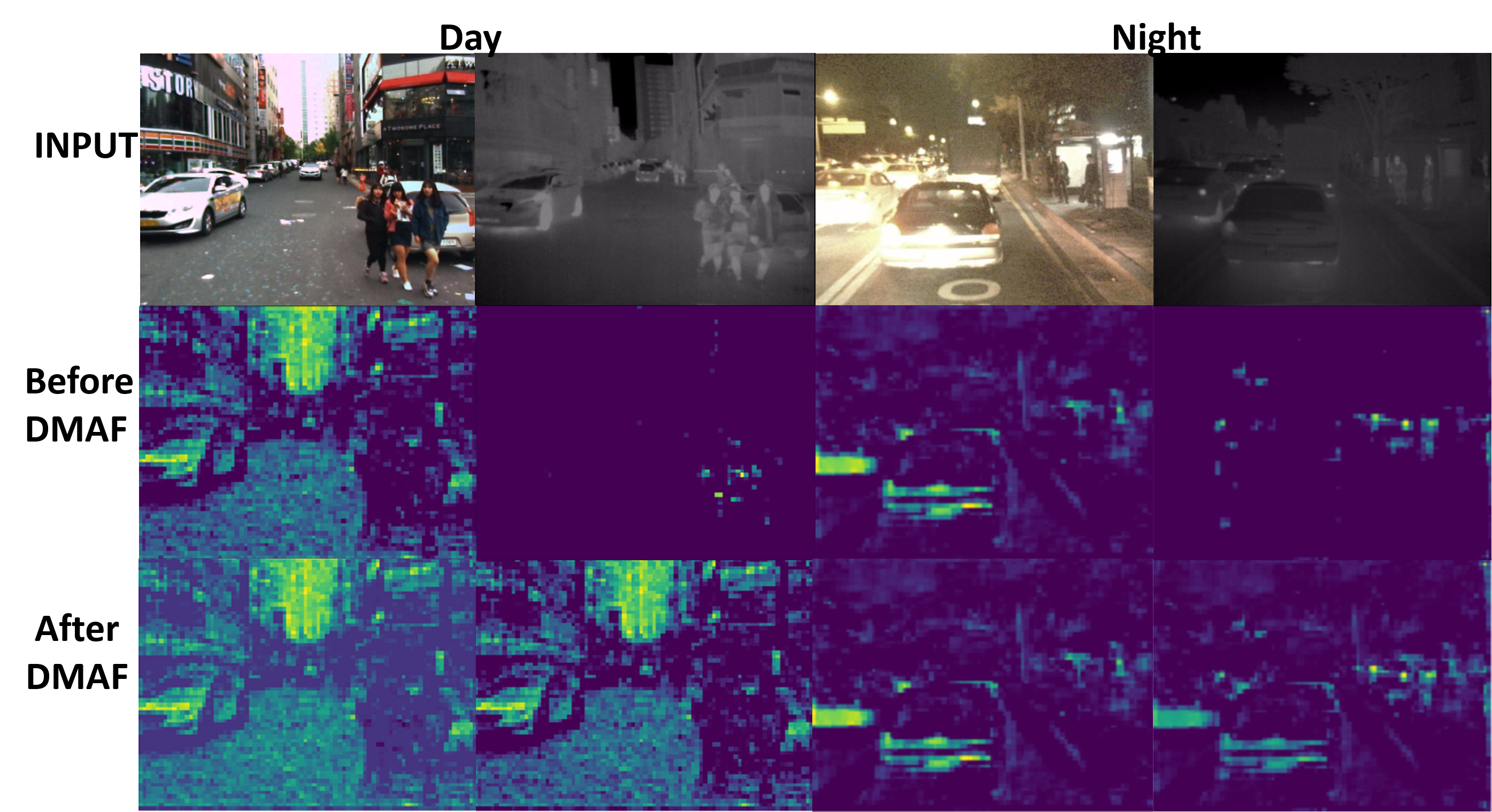}
	\end{center}
	\caption{Feature map visualization of one channel in stage3 (shown in Fig.~\ref{fig:framework}) before and after DMAF module. The two modality feature maps are remedied with the differential information from each other. }
	\label{fig:DMAF_vis}
\end{figure}

\noindent where $\mathcal{F(X)}$ is considered as the residual function. $\sigma$ refers to the tanh function, $GAP$ refers to Global Average Pooling, and $\oplus$,$\odot$ represent element-wise sum and element-wise multiplication respectively. It is noteworthy that the $F_{RD}$,$F_{TD}$ are added to the residual branch which formulates the complementary feature learning as residual learning inspired by RFBNet \cite{deng2019rfbnet}. With residual mapping, the complementary feature would not directly impact the modality-specific stream. The DMAF module acts as a part of residual function in the ResNet block.

The visualization result of DMAF module is illustrated
in Fig.~\ref{fig:DMAF_vis}. Due to the differences in the characteristics of two modalities, thermal and RGB modalities have certain limitations respectively in capturing pedestrian and background features. As the CNN goes deeper, pedestrian features gradually become salient and background features are re-integrated. The integration of background features means useful background information is refined and noisy background information is eliminated as much as possible. The DMAF module which effectively combine modality features can contribute to the integration of background information and make pedestrian features prominent from low level to high level. In our opinion, the DMAF module facilitates modality interaction in the network which reduces the learning of redundancy and conveys more information (refer the detailed analysis in appendix). In terms of no extra parameters and low computational complexity, the MBNet backbone (ResNet embedded with DMAF) may do a favor to other computer vision communities such as RGB-Depth tasks, stereo image SR, RGB-LiDAR tasks, etc.

\begin{figure}[t]
	\begin{center}
		\includegraphics[width=0.8\linewidth]{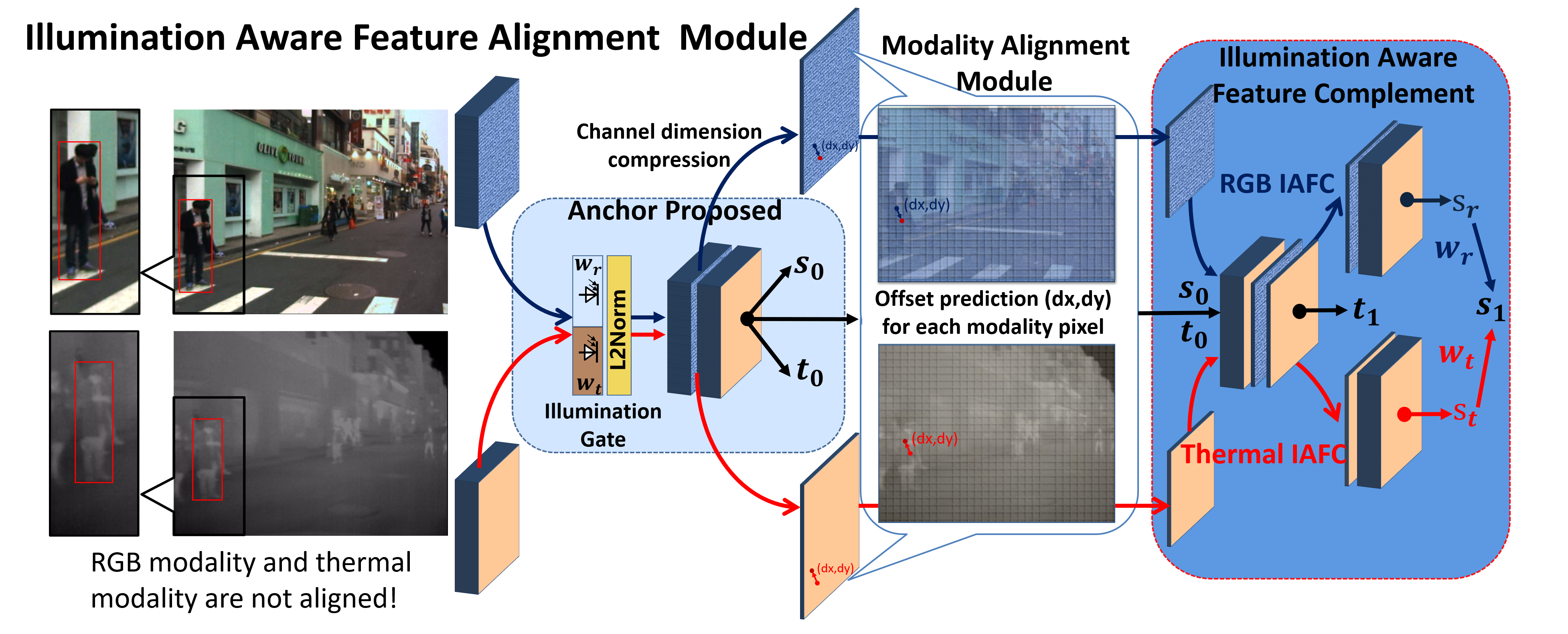}
	\end{center}
	\caption{The structure of illumination aware feature alignment module. Anchor Proposed (AP) stage generates an approximate location and Illumination Aware Feature Complement (IAFC) stage predicts based on the results of AP stage with the illumination aware balance of the two modality features. Modality Alignment (MA) module fixes the misalignments between the RGB modality and the thermal modality.}
	\label{fig:IAFA_module}
\end{figure}


\subsection{Illumination Aware Feature Alignment Module}
\label{sec:IAM} 
Illumination Aware Feature Alignment module plays the role of adapting the model to different illumination conditions and aligning the two modality features in the region proposal stage. As the top of Fig.~\ref{fig:framework} shows, we design a tiny neural network to capture the illumination values in which only the RGB images are used because the thermal images are difficult to reflect the environment illumination condition. In order to reduce computational complexity, the RGB images are resized to $56\times56$ and sent into the illumination aware module which consists of two convolutional layers and three fully-connected layers. The ReLU activation function and a $2\times2$ maxpooling layer are followed after the convolutional layer to compress and extract features. The network is optimized by minimizing cross entropy loss function between the predicted illumination values and the true labels. The illumination loss $L_{I}$ is formulated as:
\begin{equation}
\begin{split}
L_{I} &=-\widehat{w}_{d} \cdot \log \left(w_{d}\right)-\widehat{w}_{n} \cdot \log \left(w_{n}\right)  \\
w_{r} &= (  \frac{w_{d} - w_{n}}{2} ) \cdot (\alpha_{w} \cdot| w | +\gamma_{w}) + \frac{1}{2}
\ \ \ w_{t} = 1 - w_{r}
\end{split}	
\end{equation}

\noindent where $w_{d}$  and  $w_{n}$ are the softmax output of full connection layers. $\hat{w_{d}}$ and $\hat{w_{n}}$  represent the true labels of the day and night. To be self-adaptable in the network, $w_{d},w_{n}$ are readjusted in the illumination mechanism in which $|w| \in [0, 1]$ is the independent prediction of the bias from 0.5 and $\alpha_{w},\gamma_{w}$ are two learnable parameters initialized with 1, 0. Then the re-scaled results $w_{r},w_{t}$ are embedded into the network to have a balanced optimization during different illumination conditions. We tailor an illumination gate to control the weight of thermal modality stream and RGB modality stream before the Anchor Proposal (AP) stage. By element-wise multiplying with illumination value, the feature maps from two modalities have different scales after reweighting, and we use L2-normalization to rescale their norms to 10.

Considering that RGB and thermal cameras are not always captured
at the same time, there are slight misalignments between RGB and thermal modality as shown in Fig.~\ref{fig:IAFA_module}. In the fixed receptive field of a convolution kernel, the modality misalignments will cause the unbalanced feature representations and contributions of the two modalities. We contrive a Modality Alignment (MA) module which predicts offsets (dx, dy) for every pixel (x, y) of each modality. Channel dimensions are compressed  because the rearrangement of two modality features according to the learned offsets is time-consuming. Since (dx, dy) is the  float type, we adopt bilinear interpolation to obtain the final pixel value (x+dx, y+dy) from the four neighborhood pixels.

Due to the vague and sparse pedestrian distribution, we employ a cascade architecture inspired by \cite{kong2019consistent,jang2019propose,liu2018learning}. Fig.~\ref{fig:IAFA_module} shows the cascade region proposal module which consists of two stages, Anchor Propose (AP) stage and Illumination Aware Feature Complement (IAFC) stage. First, the reweighted RGB and thermal feature maps are combined to generate an approximate location estimation by AP stage. The predicted regression offsets $t_{0}$ are used to propose the deformable anchors as the basic reference for position prediction in the next IAFC stage. Then the deformable anchors and confidence scores are further fine-tuned through the IAFC stage. The confidence scores  $s_{r}$, $s_{t}$ predicted by RGB and thermal feature maps are reweighted according to illumination values. The final confidence scores  $s_{final}$ and regression offsets $t_{final}$ are computed as follows:
\begin{equation}
\begin{aligned} s_{f i n a l} &=s_{0} \times s_{1}=s_{0} \times\left(w_{r} \cdot s_{r}+w_{t} \cdot s_{t}\right) \ \ \ \ t_{f i n a l} = t_{0}+t_{1}	
\end{aligned}
\end{equation}

The multiplication in confidence scores is to encourage the final score only if two-stage scores $s_{0}, s_{1}$ are both high. While for regression offsets, summation is adopted to progressively approach the pedestrian bounding boxes. Inspired by \cite{lin2017focal}, we append the focal weight in classification loss $L_{cls}$ to address the positive-negative imbalance. The $L_{cls}$ is formulated as:
\begin{equation}
\begin{split}
L_{cls}=-\alpha \sum_{i \in S_{+}}\left(1-s_{i}\right)^{\gamma} \log \left(s_{i}\right)-(1-\alpha) \sum_{i \in S_{-}} s_{i}^{\gamma} \log \left(1-s_{i}\right)
\end{split}
\end{equation}

$S_{+}, S_{-}$ are the positive and negative anchor boxes. As suggested in \cite{lin2017focal}, we experimentally set $\alpha = 0.25$ and $\gamma = 2$. $s_{i}$ is the positive probability of samples i. The total loss is the sum of illumination loss $L_I$, classification loss $L_{cls}$ and regression loss $L_{reg}$, where the regression loss $L_{reg}$ is the smooth L1 loss raised by Faster-RCNN \cite{ren2015faster}. The total loss function $L$ is as follows:

\begin{equation}
\begin{split}
\emph{L}=L_{I}+L_{c l s 0}+L_{c l s 1}+[y=1] L_{r e g 0}+[y=1] L_{r e g 1}
\end{split}
\end{equation}

With the progressive detection of AP stage and IAFC stage, more positive cases are generated to benefit bounding box regression in the second IAFC stage. The adaptive illumination aware feature alignment of RGB modality and thermal modality provides a solution to feature modality imbalance problems by aligning two modality features, meanwhile, it also makes the detector more robust to the illumination variation.  


\begin{table*}[t]
	\caption{Comparisons with the state-of-the-art methods on the KAIST reasonable subset in terms of $MR^{-2}$ \cite{hwang2015multispectral} with different thresholds of IoU. In addition, Comparisons of running time are also provided.}
	\begin{center}
		\setlength{\tabcolsep}{0.6mm}{
			\begin{tabular}{c|c|c|c|c|c|c|c|c}
				\hline
				\multirow{2}*{Methods}  & \multicolumn{3}{c|}{$MR^{-2}$ ( IoU = 0.5 )}&\multicolumn{3}{c|}{$MR^{-2}$( IoU = 0.75 )}& \multirow{2}*{Plateform}&\multirow{2}*{Speed(s)}\\
				\cline{2-7} 
				& All & Day& Night& All & Day& Night&& \\
				\hline
				ACF \cite{hwang2015multispectral} & 47.32  &42.57& 56.17&88.79  & 87.70& 91.22& MATLAB
				& 2.73\\
				Halfway Fusion\cite{liu2016multispectral} & 25.75  & 24.88& 26.59&81.29  & 78.43& 86.80& TITAN X & 0.43\\
				Fusion RPN+BF \cite{konig2017fully} & 18.29 & 19.57& 16.27&72.97  & 68.14& 81.35& MATLAB& 0.80\\
				IAF R-CNN \cite{li2019illumination}& 15.73 & 14.55& 18.26& 75.50  & 72.34& 81.12&TITAN X& 0.21\\
				IATDNN + IASS\cite{guan2019fusion} & 14.95 & 14.67& 15.72&76.69& 76.46& 77.05& TITAN X& 0.25\\		
				RFA\cite{zhang2019cross1} & 14.61 & 16.78& 10.21&-  & -& -& TITAN X& 0.08\\
				CIAN \cite{zhang2019cross0}& 14.12 & 14.77& 11.13&74.45  & 71.42& 80.16& 1080 Ti& \textbf{0.07}\\		
				MSDS-RCNN \cite{li2018multispectral}& 11.34 & 10.53& 12.94&70.57 &67.36 & 79.25& TITAN X& 0.22\\
				AR-CNN \cite{zhang2019weakly} & 9.34 & 9.94& 8.38& 64.22 & 57.87& 76.82& 1080 Ti&0.12\\
				MBNet(ours) &\textbf{8.13} & \textbf{8.28}& \textbf{7.86}&\textbf{60.12} & \textbf{54.90}& \textbf{68.34}& 1080 Ti& \textbf{0.07}\\
				\hline		
		\end{tabular}}
	\end{center}
	\label{MR_speed}
\end{table*}

\section{Experiments}

In this section, we first introduce the KAIST dataset \cite{hwang2015multispectral} and CVC-14 dataset \cite{gonzalez2016pedestrian}. Then we show implementation details and experiment results to compare logMR and runtime of the proposed MBNet with the state-of-the-art methods. The evaluation is based on the reasonable setup  (55 pixel or taller under partial or no occlusion) unless otherwise mentioned. Finally, we will carry out ablation studies for the proposed method on the KAIST dataset.

\subsection{Datasets}
Our approach is evaluated on the KAIST dataset \cite{hwang2015multispectral} and CVC-14 dataset \cite{gonzalez2016pedestrian}.

\textbf{KAIST}. The KAIST dataset \cite{hwang2015multispectral} contains 95,328 aligned color-thermal image pairs, with a total of 103,128 bounding boxes covering 1,182 unique pedestrians. Due to the problematic annotations in original training data, we adopt the  annotations improved by Zhang et al. \cite{zhang2019weakly} for training. The test set consists of 2, 252 frames sampled every 20th frame from video, among which 1,455 images are captured during daytime and the rest 797 images are during nighttime. The evaluation metric follows the standard KAIST evaluation \cite{hwang2015multispectral}: log-average Miss Rate over False Positive Per Image (FPPI) range of [$10^{-2},10^{0}$] (denoted as $MR^{-2}$). We evaluate the detection performance on the KAIST test set with annotations improved by Liu et al. \cite{liu2016multispectral} and report the runtime of the proposed MBNet using a single NVIDIA GTX 1080Ti GPU for fair comparison with the state-of-the-art methods before. 

\textbf{CVC-14}. The CVC-14 dataset \cite{gonzalez2016pedestrian} contains visible (grayscale) and thermal paired images. It was recorded in various scenes at day and night by on-board color and thermal cameras at 10 Hz. The training and testing set contains 7, 085 and 1, 433 frames, respectively. Annotations are individually provided in each modality since the cameras are not well calibrated.

\subsection{Implementation Details}

Our MBNet detector uses ResNet-50 \cite{He_2016_CVPR} as the backbone network, which is pretrained on ImageNet unless otherwise stated. The training IoU of the AP stage is set to \{0.3, 0.5\} and the IAFC stage is set to {\{0.5, 0.7\}}.  The Xavier method \cite{glorot2010understanding}  is used to randomly initialize other convolutional layers. In the training, we crop a patch with the size of [0.3, 1] of the input image and resize it to $640 \times 512$, then each image is randomly color distorted and horizontal flipped with a probability of 0.5 to increase the diversity. The whole network is trained by adam optimizer for 7 epoches with a learning rate of 0.0001 and a batch size of 10. Followed by \cite{li2018multispectral}, the width of initial anchors are set to [25.84, 29.39], [33.81, 38.99], [44.47, 52.54], [65.80, 131.40] for stage3 to stage6 ( shown in Fig. \ref{fig:framework} ) with a single anchor ratio of 0.41.

\begin{table}[t]
	\caption{Evaluations on the KAIST dataset under all nine test subsets and ablation experiments are also provided. The lowest MR are highlighted with bold font. }
	\begin{center}
		
		\setlength{\tabcolsep}{0.7mm}{
			\begin{tabular}{ ccc |c|c|c|c|c|c|c|c|c}
				\hline
				\multicolumn{3}{c|}{Methods} & Rea. & Day& Night& Near & Medium& Far&None&Partial&Heavy \\
				\hline
				\multicolumn{3}{c|}{ACF \cite{hwang2015multispectral}} & 47.32  &42.57& 56.17&28.74  & 53.67& 88.20& 62.94	& 81.40&88.08\\
				\multicolumn{3}{c|}{Halfway Fusion \cite{liu2016multispectral} }& 25.75  & 24.88& 26.59&8.13&30.34&75.70 & 43.13&65.21 & 74.36\\
				\multicolumn{3}{c|}{FusionRPN+BF \cite{konig2017fully}} & 18.29 & 19.57& 16.27&0.04  & 30.87& 88.86& 47.45& 56.10&72.20\\
				\multicolumn{3}{c|}{IAF R-CNN \cite{li2019illumination}}& 15.73 & 14.55& 18.26&0.96  & 25.54& 77.84&40.17& 48.40&69.76\\
				\multicolumn{3}{c|}{IATDNN+IASS \cite{guan2019fusion}} & 14.95 & 14.67& 15.72&0.04  & 28.55& 83.42&45.43& 46.25&64.57\\
				\multicolumn{3}{c|}{CIAN \cite{zhang2019cross0}}& 14.12 & 14.77& 11.13&3.71 & 19.04& 55.82& 30.31& 41.57&62.48\\		
				\multicolumn{3}{c|}{MSDS-RCNN \cite{li2018multispectral}}& 11.63 & 10.60& 13.73&1.29 &16.19 & 63.73& 29.86& 38.71&63.37\\
				\multicolumn{3}{c|}{AR-CNN \cite{zhang2019weakly} }& 9.34 & 9.94& 8.38& 0.00 & 16.08& 69.00&31.40&38.63&\textbf{55.73}\\
				\hline
				\multicolumn{12}{c}{MBNet (ours) }\\
				\hline			
				IAFC  & DMAF &Aligned    \\ 
				\cline{1-12} 
				& &&  11.93&12.51&10.86&0.00 &20.08& 63.70& 34.16  & 44.10&65.11 \\
				\ \  & \ \ \ $\checkmark$ && 10.96 & 11.13  & 10.48&0.00 & 20.33 &61.25 &33.50 & 39.80  & 62.68  \\
				\ \ $\checkmark$   & &\ \ \ &10.53 &11.00& 9.75& 0.00  &16.50& 58.47& 29.39  & 40.25& 59.13\\
				\ \ $\checkmark$ &\ \ \ $\checkmark$&  & 9.36 & 9.72& 8.63 & 0.00  & 16.18& \textbf{54.66}& 28.02  & 38.19& 60.70\\
				\ \ $\checkmark$ &\ \ \ $\checkmark$ &\ \ \ $\checkmark$ & \textbf{8.13}& \textbf{8.28}& \textbf{7.86}& \textbf{0.00}& \textbf{16.07}& 55.99  & \textbf{27.74}& \textbf{35.43}& 59.14  \\
				
				\hline
		\end{tabular}}
	\end{center}
	
	\label{alltest_set}
\end{table}

\subsection{Evaluation on the KAIST Dataset}
We show the superiority of our method from both aspects of miss rate ( MR ) and speed.

\textbf{Miss Rate.} Our proposed approach achieves 8.28 MR, 7.86 MR, and 8.13 MR on the reasonable day, night and all-day subset respectively under the IoU threshold of 0.5, all of them are lower than the previous best competitor AR-CNN \cite{zhang2019weakly}. In the case of a stricter IoU threshold of 0.75, Tab.~\ref{MR_speed} shows our proposed method achieves about 4.10\% lower on $MR^{-2}$ which implies that the MBNet has a substantially better localization accuracy compared with AR-CNN. The larger the IoU threshold is set, the harder the predicted bounding boxes are considered to be True Positives (TP) in the evaluation. In order to have a comprehensive understanding of detector performance, we also make an evaluation under all nine subsets including the pedestrian distances and the occlusion levels. Tab.~\ref{alltest_set} shows MBNet outperforms other methods under most subsets with no extra treatment to handle the small and occlusion pedestrians, especially on the none subset (27.74 vs. 29.86 MR) and partial subset (35.43 vs. 38.63 MR). 

\begin{figure}[htbp]	
	\centering
	\includegraphics[width=0.85\linewidth]{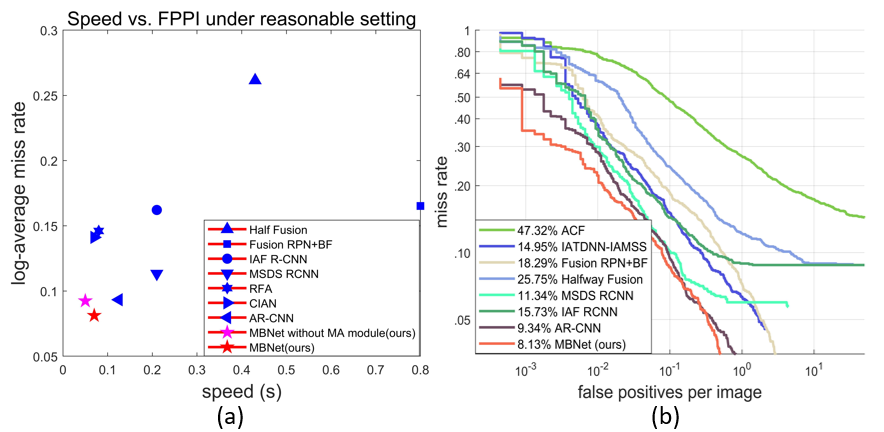}			
	\caption{(a) Log-average miss rate versus the running time of each detector. (b) Performance comparisons with the state-of-the-art methods on the KAIST dataset under reasonable subset. }		
	\label{fig:speed_FPPI}	
\end{figure}

\textbf{Speed.} We also compare the running time of MBNet with state-of-the-art methods. MBNet directly takes 640 $\times$ 512 multispectral images as input without image up-scaling. Since modality alignment module is time-consuming, we draw the Speed vs. FPPI results of MBNet and MBNet without MA module respectively in Fig. \ref{fig:speed_FPPI} (a). The MBNet without MA module reaches the fastest speed of 20 fps and has a comparable performance with AR-CNN \cite{zhang2019weakly}. MBNet achieves the state-of-art performance on the test annotations improved by Liu et al. \cite{liu2016multispectral} while maintaining high computational efficiency. 

Overall, the Speed vs. FPPI and performance comparisons under reasonable setting are shown in Fig.~\ref{fig:speed_FPPI} (a) and (b). The result indicates that MBNet is an attractive multispectral pedestrian detector in both accuracy and speed. 

\subsection{Evaluation on the CVC-14 Dataset}
We fine-tune from the KAIST pretrained model in the training of CVC-14 dataset. We follow the protocol in \cite{park2018unified} to conduct the evaluation experiments and adopt the strategy in \cite{zhang2019weakly}. Specifically, there exsiting serious misalignments between thermal and RGB modalities, so we consider pedestrians in the RGB modality as the training target, and pedestrians in the thermal modality act as a reference. It can be observed from Tab.~\ref{CVC-14}. that MBNet can still achieve good results even in the case of serious modality misalignments, which demonstrates
that modality balance strategy improves the robustness to the position shift problem.



\begin{table}[t]
	\caption{Evaluation results on the CVC-14 dataset. The first column refers to input modalities of the approach. }
	\begin{center}
		\setlength{\tabcolsep}{1.1mm}{
			\begin{tabular}{c|c|c|c|c|c|c|c|c|c c}
				\cline{1-10}
				\multirow{8}{*}{\rotatebox{90}{Visible}} & \multirow{2}{*}{Methods} & \multicolumn{3}{c|}{$MR^{-2}$} & \multirow{8}{*}{\begin{tabular}[c]{@{}c@{}}\rotatebox{90}{Thermal}\\ \rotatebox{90}{Visible+}\end{tabular}} & \multirow{2}{*}{Methods} & \multicolumn{3}{c}{$MR^{-2}$} & \multicolumn{1}{l}{} \\ \cline{3-5} \cline{8-10}
				&                         & Day   & Night   & All   &                                                                            &                         & Day   & Night   & All   
				\\ \cline{2-5} \cline{7-10}
				& SVM  \cite{gonzalez2016pedestrian}                     & 37.6     & 76.9       & -     &                                                                            & MACF \cite{park2018unified}                      & 61.3    & 48.2       & 60.1     &                      \\
				& DPM \cite{gonzalez2016pedestrian}                        & 25.2    & 76.4      & -     &                                                                            & Choi et al. \cite{choi2016multi}                       & 49.3     & 43.8       & 47.3     &                      \\ 
				& Random Forest \cite{gonzalez2016pedestrian}                       & 26.6    & 81.2       & -    &                                                                            & Halfway Fusion \cite{park2018unified}                     & 38.1    & 34.4      & 37.0    &                      \\ 
				& ACF \cite{park2018unified}   & 65.0    & 83.2      & 71.3    &           &   Park et al.  \cite{park2018unified}    & 31.8    & 30.8      & 31.4    &						 \\
				& Faster R-CNN \cite{park2018unified}                     & 43.2    & 71.4      & 51.9    &                                                                            & AR-CNN \cite{zhang2019weakly}                     & 24.7    & 18.1     & 22.1    & 				 \\
				&                        &     &       &     &                                                                            & MBNet （(ours)                     & \textbf{24.7}   &  \textbf{13.5}      & \textbf{21.1}    & 				 \\
				
				\cline{1-10}
			\end{tabular}
		}
	\end{center}
	\label{CVC-14}
\end{table}

\subsection{Ablation Study}
Ablation experiments are performed on the KAIST dataset for a detailed analys in this section. The baseline is initialized from two-stream SSD \cite{liu2016ssd} and adopts simple concatenation to fuse two modalities as most previous methods do. We show how to construct the MBNet with the principle of modality balance.

\textbf{Differential Modality Aware Fusion.}
Simple concatenation modulates the complementary part implicitly and has a large computational cost. It encourages us to seek a better integration of modalities. DMAF modules are inserted in each ResNet block and act as a part of residual function to be learned which will impel a deep integration of RGB and thermal modalities. From the ablation experiment results in  Tab.~\ref{alltest_set} , we could see that the MBNet generates more accurate detection results with DMAF module added. The DMAF module capture the complementary modality information in a more explicit way with no extra parameters and very little computation, so that the MBNet can maintain a fast speed. The feature representation can be more balanced after the two modalities are fully integrated.

\textbf{Modality Alignment.} It could be seen from the Tab.~\ref{IAFA} that the MBNet with modality alignment module has a much lower miss rate on the reasonable subset. It indicates that the MA module with bilinear interpolation used can locate pedestrians more precisely especially in the deeper network which has a smaller resolution of the feature maps. After the MA module, the RGB and thermal modality features are rearranged and the aligned features can make a balanced contribution at one position. In addition, it can be observed that MA module narrows the gap of miss rates between day and night subsets. The experiment results demonstrate that the illumination modality imbalance problem will be mitigated with the alignment of feature modalities. The illumination and feature modality imbalance problems exist side by side and play a part together. 

\textbf{Illumination Aware Feature Alignment.} To have a deep insight into the effectiveness of illumination aware feature alignment module, we investigate the performance of different design choices in Tab.~\ref{IAFA}. RGB IAFC and thermal IAFC represent complementing the prediction results of AP stage with RGB and thermal modality features according to the illumination conditions. It is observed that performance gains can generally be achieved by illumination aware feature complement. RGB modality is beneficial to the pedestrian detection during the day while thermal modality is beneficial to the night. By introducing the illumination gate which applies weights to the RGB and thermal stream, MBNet has a more balanced performance under different illumination conditions. This demonstrates that the detection performance can be further improved by illumination aware mechanism, since it helps the network mitigate illumination modality imbalance problem.

\begin{table}[t]
	\caption{Ablation experiments of illumination aware feature alignment module evaluated on KAIST reasonable set. }
	\begin{center}
		\setlength{\tabcolsep}{2.8mm}{
			\begin{tabular}{c|c|lllll}
				\hline
				\multicolumn{2}{c|}{Component}&\multicolumn{4}{c}{Choice}\\
				\hline
				\multicolumn{2}{c|}{RGB IAFC}&$\checkmark$& &$\checkmark$&$\checkmark$&$\checkmark$\\
				\multicolumn{2}{c|}{Thermal IAFC}& &$\checkmark$&$\checkmark$&$\checkmark$&$\checkmark$\\
				\multicolumn{2}{c|}{Illumination Gate}&$\checkmark$&$\checkmark$& &$\checkmark$&$\checkmark$\\
				\multicolumn{2}{c|}{Modality Alignment}&&& &&$\checkmark$\\
				\hline
				\multirow{3}*{MR(\%)} &Day&9.80&  10.05 &10.74 &9.72 &8.28\\
				&Night&10.17& 9.76  &9.54 &8.63 &7.86\\
				&All&9.86& 9.89   &10.27 &9.36 &8.13\\
				\hline
		\end{tabular}}
	\end{center}
	
	\label{IAFA}
\end{table}

\section{Conclusion}

In this work, we explore a one-stage detector named MBNet to alleviate the modality imbalance problems in multispectral pedestrian detection. Specifically, the DMAF module is densely inserted in the ResNet block to fully integrate features and the MA module aligns two modalities so that the RGB and thermal features can have an equal contribution and representation. Meanwhile, the illumination gate embedded in the backbone network and the adaptive illumination aware feature complement in the region proposal stage make the detector robust to the variant illumination. We argue the modality imbalance problems are not limited to multispectral pedestrian detection. They are widespread in multimodal computer vision task to which the balance and integration of different modality features should be paid attention. We will further study how to reconcile the balance and reduce the learning of redundancy between different modalities in other computer vision task in the future.

\noindent\textbf{Acknowledgment}. This work was supported in part by the National Natural Science Foundation of China (Nos. 61627804).

\clearpage
%
%
\bibliographystyle{splncs04}
\bibliography{egbib}

\clearpage
\section{Appendix}

We further clarify the meaning of $F_{D}$ here. For the thermal modality, $F_{D} = F_{R} - F_{T}$. for the RGB modality, $F_{D} = F_{T} - F_{R}$. So the Eq. 2 can be rewritten as follows:
$$
F_{T}^{\prime} 
= F_{T}+\mathcal{F}\left(F_{T} \oplus\left(\sigma\left(G A P\left(F_{D}\right)\right) \odot F_{R}\right)\right) = F_{T}+\mathcal{F}\left(F_{T} \oplus\left(\sigma\left(G A P\left(F_{R}-F_{T}\right)\right) \odot F_{R}\right)\right)
$$
$$
F_{R}^{\prime} = F_{R}+\mathcal{F}\left(F_{R} \oplus\left(\sigma\left(G A P\left(F_{D}\right)\right) \odot F_{T}\right)\right)
= F_{R}+\mathcal{F}\left(F_{R} \oplus\left(\sigma\left(G A P\left(F_{T}-F_{R}\right)\right) \odot F_{T}\right)\right)
$$

\begin{figure}[h]
	\begin{center}
		\includegraphics[width=1.0\linewidth]{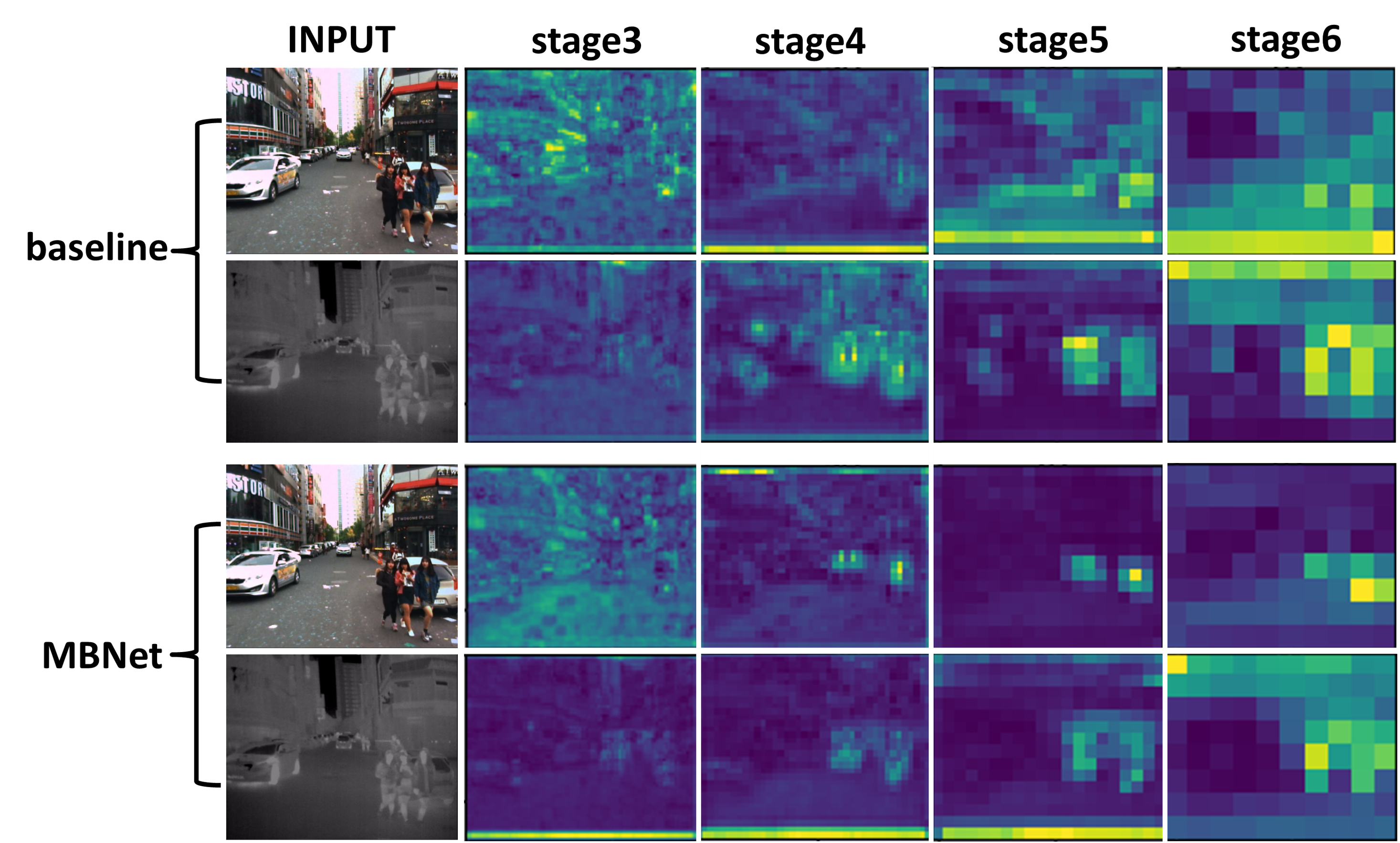}
	\end{center}
	\caption{ The overall perspective of the baseline and MBNet feature maps ($H\times W\times C$)  in stage3, stage4, stage5, stage6. }
	\label{fig:DMAF_all}
\end{figure}

\begin{figure}[t]
	\begin{center}
		\includegraphics[width=1.0\linewidth]{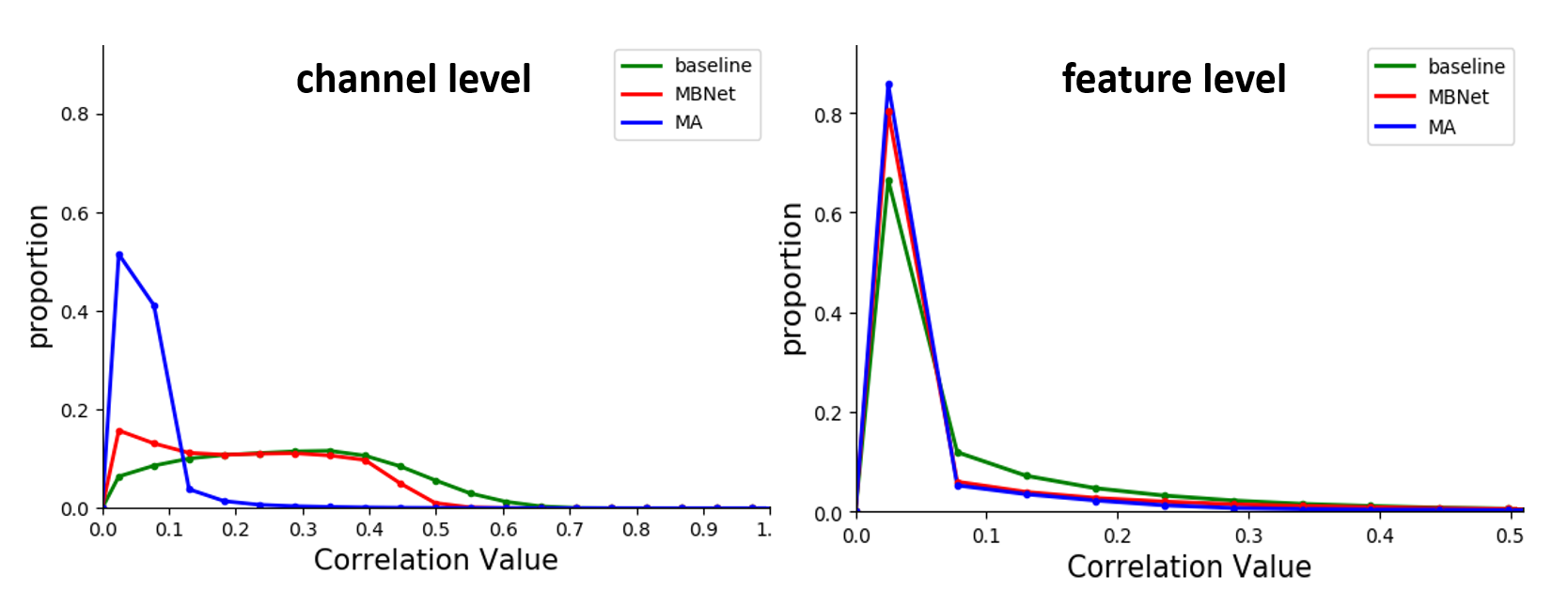}
	\end{center}
	\caption{The modality redundancy (Pearson product-moment correlation coefficient $|\rho|$) between two modality feature maps from channel level  ($1\times 1\times C$) and feature level ($H\times W\times 1$). The green and red line represent the baseline and MBNet stage3--6 feature maps respectively. The blue line represents the feature maps after Modality Alignment (MA) module in MBNet.   }
	\label{fig:correlation}
\end{figure}

The DMAF module fuses two modalities at the channel level, so it should be noted that the visualization in Fig.~\ref{fig:DMAF_vis} is just one channel ($H\times W\times 1$) in order to have an intuitive understanding of the effect of DMAF. From an overall perspective of the feature maps ($H\times W\times C$) shown in Fig.~\ref{fig:DMAF_all}, the pedestrian region features become more salient with DMAF added. In order to have a deeper understanding of the DMAF module, we try to explain the role of the DMAF  module from the view of modality redundancy.

we introduce the Pearson product-moment correlation coefficient ($|\rho|$) between two modality feature maps to represent the modality redundancy. Since the feature maps are three-dimensional data ($H\times W\times C$), we divide the feature maps into channel level ($1\times 1\times C$) and feature level ($H\times W\times 1$).  We randomly select 100 pair-images from the KAIST test set and the Pearson product-moment correlation coefficient ($|\rho|$) is calculated from both levels. The stage3, stage4, stage5, stage6 feature maps in the backbone (which are used to predicts the location and score) from the baseline and MBNet as well as feature maps after modality alignment module are chosen as the experimental samples. We make statistics on the correlation coefficients $|\rho|$ between two modalities and illustrate the proportion of different correlation coefficients $|\rho|$ (0.0$ \sim $1.0) as a line chart in Fig.~\ref{fig:correlation}.

 With the DMAF module added, two modalities tend to be unrelated from both the channel and feature level (red line vs. green line), which means redundant information is reduced. After the modality alignment module, the correlation between the two modalities is further reduced, especially at the channel level.   In our opinion, \textbf{the DMAF module facilitates modality interaction in the network which reduces the learning of redundancy and conveys more information}.  
The effective extraction of useful information and the elimination of redundancy between two modalities are problems worthy of studying in the future.

\end{document}